\documentclass[conference]{IEEEtran}

\makeatletter % changes the catcode of @ to 11
\newcommand{\linebreakand}{%
  \end{@IEEEauthorhalign}
  \hfill\mbox{}\par
  \mbox{}\hfill\begin{@IEEEauthorhalign}
}
\makeatother % changes the catcode of @ back to 12

\IEEEoverridecommandlockouts
\usepackage{cite}
\usepackage{amsmath,amssymb,amsfonts}
\usepackage{algorithmic}
\usepackage{graphicx}
\usepackage{textcomp}
\usepackage{xcolor}
\def\BibTeX{{\rm B\kern-.05em{\sc i\kern-.025em b}\kern-.08em
    T\kern-.1667em\lower.7ex\hbox{E}\kern-.125emX}}
\begin{document}

\title{Pushing Boundaries: Exploring Zero Shot Object Classification with Large Multimodal Models}

\author{\IEEEauthorblockN{Ashhadul Islam}
\IEEEauthorblockA{\textit{College of Science and Engineering} \\
\textit{Hamad Bin Khalifa University}\\
Doha, Qatar \\
0000-0002-9717-3252}
\and
\IEEEauthorblockN{Md. Rafiul Biswas}
\IEEEauthorblockA{\textit{College of Science and Engineering} \\
\textit{Hamad Bin Khalifa University}\\
Doha, Qatar \\
0000-0002-5145-1990}
\and
\IEEEauthorblockN{Wajdi Zaghouani}
\IEEEauthorblockA{\textit{College of Humanities and Social Sciences} \\
\textit{Hamad Bin Khalifa University}\\
Doha, Qatar \\
0000-0003-1521-5568}

\linebreakand % <----- NOTE HERE, breaking after the third one!

\IEEEauthorblockN{Samir Brahim Belhaouari}
\IEEEauthorblockA{\textit{College of Science and Engineering} \\
\textit{Hamad Bin Khalifa University}\\
Doha, Qatar \\
0000-0003-2336-0490}

\and
\IEEEauthorblockN{Zubair Shah}
\IEEEauthorblockA{\textit{College of Science and Engineering} \\
\textit{Hamad Bin Khalifa University}\\
Doha, Qatar \\
0000-0001-7389-3274}
}

\maketitle

\begin{abstract}
The synergy of language and vision models has given rise to Large Language and Vision Assistant models (LLVAs), designed to engage users in rich conversational experiences intertwined with image-based queries. These comprehensive multimodal models seamlessly integrate vision encoders with Large Language Models (LLMs), expanding their applications in general-purpose language and visual comprehension. The advent of Large Multimodal Models (LMMs) heralds a new era in Artificial Intelligence (AI) assistance, extending the horizons of AI utilization. This paper takes a unique perspective on LMMs, exploring their efficacy in performing image classification tasks using tailored prompts designed for specific datasets. We also investigate the LLVAs zero-shot learning capabilities. Our study includes a benchmarking analysis across four diverse datasets: MNIST, Cats Vs. Dogs, Hymnoptera (Ants Vs. Bees), and an unconventional dataset comprising Pox Vs. Non-Pox skin images. The results of our experiments demonstrate the model's remarkable performance, achieving classification accuracies of 85\%, 100\%, 77\%, and 79\% for the respective datasets without any fine-tuning. To bolster our analysis, we assess the model's performance post fine-tuning for specific tasks. In one instance, fine-tuning is conducted over a dataset comprising images of faces of children with and without autism. Prior to fine-tuning, the model demonstrated a test accuracy of 55\%, which significantly improved to 83\% post fine-tuning. These results, coupled with our prior findings, underscore the transformative potential of LLVAs and their versatile applications in real-world scenarios.
\end{abstract}


\section{Introduction}
An image chatbot represents a unique form of interactive Artificial Intelligence (AI) system explicitly engineered to comprehend and respond to user inputs that encompass visual content, setting it apart from traditional text-based chatbots. Unlike its text-centric counterparts, an image chatbot possesses the capability to scrutinize visual data, enabling it to furnish responses that are not only contextually relevant but also exceptionally precise, as previously noted \cite{PATIL2023}. Users are empowered to submit images, either through direct uploads or image-sharing within the chat interface, and the chatbot harnesses its image analysis proficiency to address inquiries, provide clarifications, and even offer recommendations, all rooted in the visual cues contained within the images. This groundbreaking technology has found extensive utility across a multitude of domains, from revolutionizing customer service and enhancing healthcare to elevating the retail experience. It augments user engagement and emotions by introducing a novel and compelling way for individuals to interact with AI systems \cite{info:doi/10.2196/48659, lee2023building}.

The fusion of chatbot technology with the domain of medical image analysis represents a particularly notable development as researchers actively explore methods to incorporate image acquisition from within the human body. This innovation enables the creation of visual representations that are invaluable in the context of clinical decision-making and medical interventions \cite{HANDA2023105292}. The integration of image processing capabilities into chatbots constitutes a pivotal advancement, furnishing users with conversational agents that are not just more intelligent but also substantially more functional, accessible, and engaging. This progress towards creating dynamic and interactive user interactions reflects the potential of chatbots to revolutionize the way people interact with AI technology \cite{HANDA2023105292}.

Classifier algorithms, on the other hand, function as analytical tools that precisely analyze input data, identify distinguishing features associated with various categories and then assign data to specific categories based on their inherent characteristics. Well-established classifier algorithms include Decision Trees, Random Forest, Support Vector Machines (SVM), Naive Bayes, and Logistic Regression. These classifiers generally undergo training with labeled datasets, learning to establish associations between particular input features and corresponding categories. Advanced multimodal large language models (LLMs) stand out as an exemplary class of models capable of generating responses by seamlessly integrating diverse forms of information, including images, text, and audio files \cite{bagdasaryan2023abusing}.

The fusion of image chatbots with classifier algorithms directs in a new dimension of functionality, where the chatbot's capacity to analyze visual data converges with the classifier's ability to ascertain contextual intent. Consequently, the chatbot leverages this analysis to discern the context or purpose behind a user's inquiry, thereby facilitating the generation of pertinent and precise responses. This synergy between image chatbots and classifiers stands to enhance user experiences and satisfaction by providing responses that are not just relevant but deeply attuned to the specific context of the query.

\subsection{Contributions}

This paper investigates using the LLaVA 1.5 Large multimodal model for image classification datasets. The two main contributions of the paper include:

\begin{itemize}
    \item Benchmarking the model's versatility, repurposing it from image interpretation and conversation to building classifiers and extending its use to medical datasets. Performing zero shot classification of images using prompt engineering
    \item Investigating further enhancements to the model, we focus on performance improvement through fine-tuning. By refining the model's parameters and adapting it to specific tasks, we aim to elevate its overall effectiveness and applicability.
\end{itemize}

\section{Large Language and Vision Assistant (LLaVA)}
To gain a comprehensive understanding of the model under examination in this paper, LLaVA 1.5, it is essential to first familiarize ourselves with its precursor, LLaVA \cite{liu2023visual}. LLaVA is widely recognized for its adeptness in tasks related to visual reasoning. It excels in practical visual instruction-following benchmarks, although it faces limitations in academic benchmarks that demand succinct responses, primarily due to its absence of extensive pretraining. Remarkably, LLaVA demonstrates exceptional data efficiency, surpassing alternative methods while consuming fewer computational resources and requiring less training data \cite{liu2023visual}.
\subsection{Components of LLaVA}
While building the model, the researchers have established a connection between the pre-trained CLIP ViT-L/14 \cite{radford2021learning} visual encoder and the large language model LLAMA \cite{alpaca}. The visual encoder's function is to capture visual attributes from input images and link them with language embeddings via a trainable projection matrix. This projection matrix essentially serves as a conduit, converting visual features into language embedding tokens, thus facilitating the seamless integration of text and images. LLAMA is then utilized to answer the questions pertaining to the image. The instruction-tuning process encompasses two stages:

\begin{itemize}
    \item In the first stage, known as "Pre-training for Feature Alignment," updates are limited to the projection matrix, which is based on a subset of CC3M.
    \item The second stage, denoted as "Fine-tuning End-to-End," involves the simultaneous updating of both the projection matrix and the LLM. This fine-tuning occurs in two distinct usage scenarios:
    \begin{itemize}
        \item Visual Chat: LLaVA undergoes fine-tuning using generated multimodal instruction-following data tailored for user-centric daily applications.
        \item Science QA: LLaVA receives fine-tuning on a specialized multimodal reasoning dataset designed for scientific domain applications.

    \end{itemize}

\end{itemize}

\begin{figure}[htbp]
    \centering
        \fbox{\includegraphics[scale=.55]{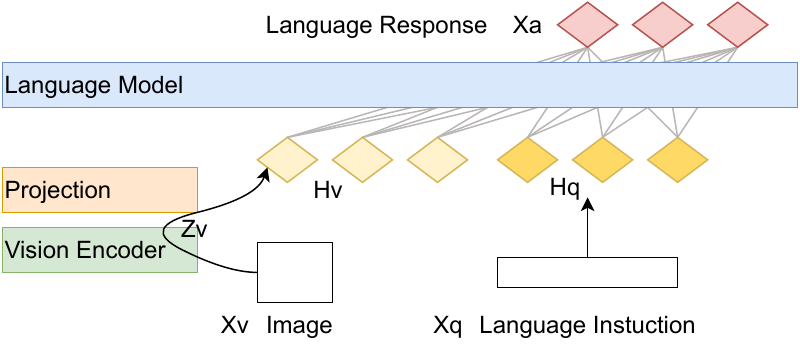}}
    \caption{Architecture of LLaVA}
    \label{FIG:LLAVA15}
\end{figure}

The authors have unfolded the inherent challenges associated with achieving a balance between short and long-form answers in visual question-answering and have proposed an innovative solution involving response formatting prompts \cite{liu2023visual}. By fine-tuning LLaVA with these prompts, which unambiguously specify the desired output format, the model becomes highly adaptable to user instructions, obviating the need for additional data processing. This approach significantly amplifies LLaVA's performance, particularly when incorporating VQAv2 data \cite{liu2023visual}.

\subsection{LLaVA 1.5}
The subsequent exploration of LLaVA 1.5 \cite{liu2023improved} entails an investigation into the consequences of transitioning from a linear projection to a two-layer MLP for the vision-language connector, leading to an augmentation of LLaVA's multimodal capabilities. Moreover, this study introduces additional academic-task-oriented datasets to enhance the model's proficiencies. Scaling is further achieved through the augmentation of input image resolution, the inclusion of GQA data \cite{hudson2018gqa}, and the expansion of the LLM size to 13B. The culminating iteration, known as LLaVA-1.5, markedly outperforms the original LLaVA, particularly in the MM-Vet benchmark, underscoring the pivotal role of the base LLM's capacity in the realm of visual conversations. According to the authors \cite{liu2023improved}, LLaVA-1.5 attains peak performance with a straightforward architecture, accessible computational resources, and publicly available datasets, establishing a fully replicable and cost-effective foundation for forthcoming research endeavors.

\subsection{LLaVA1.5 in action}

As illustrated in Table \ref{TBL:InputExample}, the model's responses vary depending on the specific prompts when provided with the same image. It demonstrates its ability to provide both detailed, extensive descriptions of the image's components and more concise answers. When tasked with image description, the model delivers a precise breakdown of various elements within the image while incorporating its own inferred insights. On the other hand, when presented with a succinct inquiry, such as counting the objects in the image, the model adeptly furnishes a straightforward response, delivering only the numerical count of objects featured in the image. This versatility in response generation showcases the model's adaptability to different types of queries and its capacity to offer relevant and context-appropriate answers. In our experiment, we leverage this capability of providing succinct answers for classification tasks.

\begin{table} [htb]
\caption{Analyzing the Impact of Various Prompts on Output Format Regularization}
\begin{center}
\begin{tabular}{|p{0.75in}p{2in}|}\hline

&\includegraphics[width=0.25\textwidth, height=40mm]{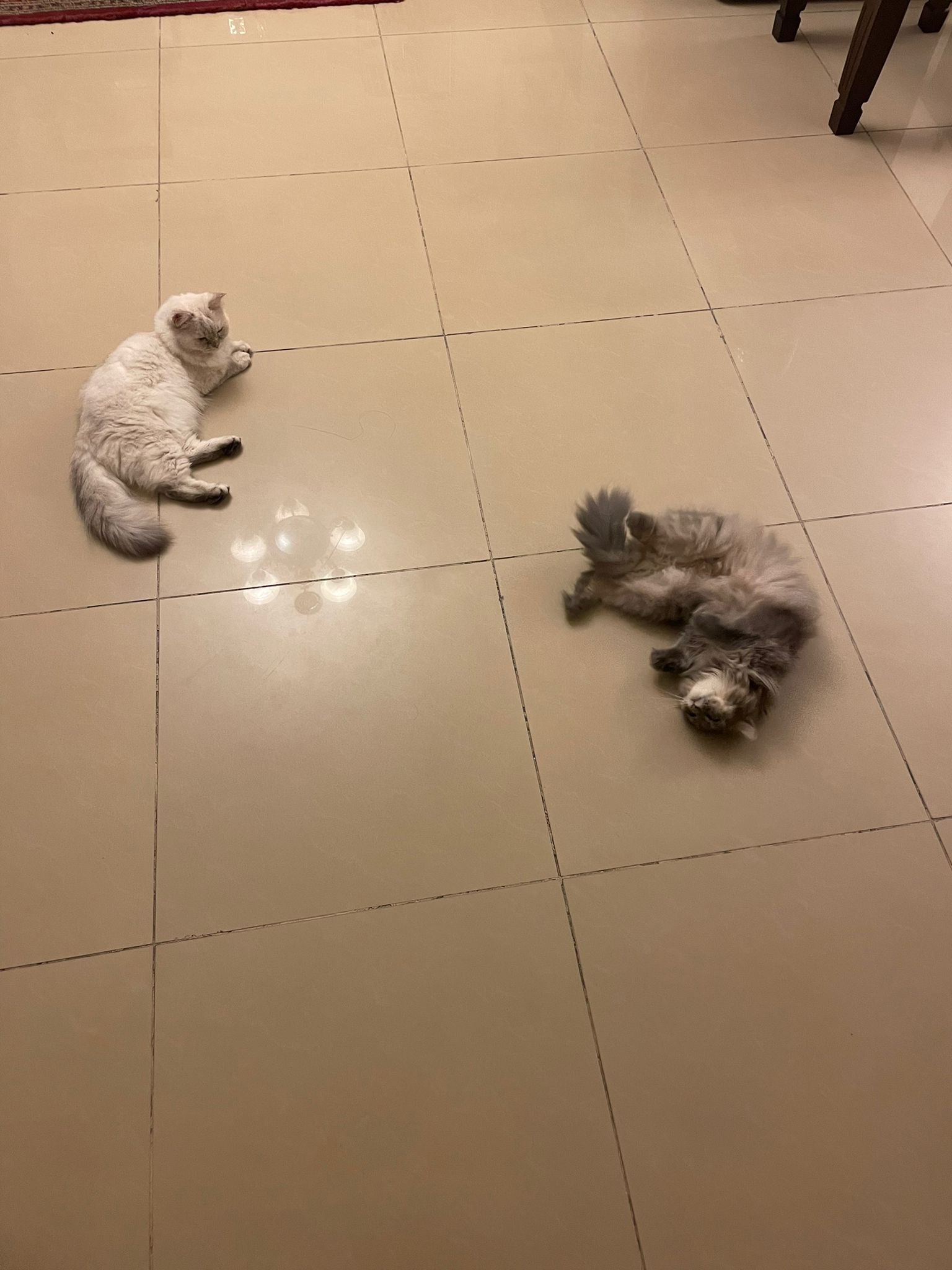}\\

Normal Prompt&Describe the image.\\
Response&The image features two cats lying on the floor, one on the left side and the other on the right side of the room. They are both resting comfortably on a tiled floor, possibly in a kitchen or dining area. The cats are positioned close to each other, creating a sense of companionship between them.\\
\hline
Formatting prompt&How many cats are present in the image. Give only the number, nothing else.\\
Response&2\\
\hline
\end{tabular}
\end{center}
\label{TBL:InputExample}
\end{table}

\section{Methodology}
\begin{figure}[htbp]
    \centering
        \fbox{\includegraphics[scale=.4]{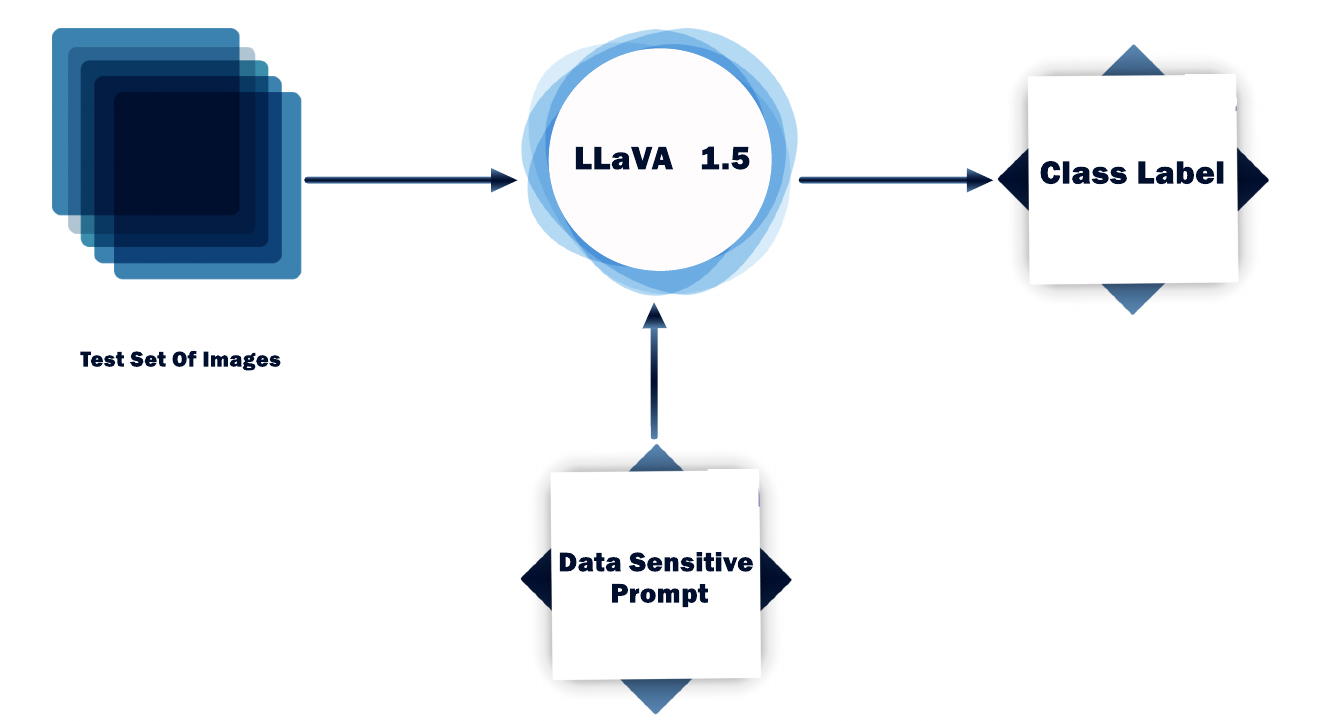}}
    \caption{Overall methodology of the experiment. The class label was achieved using a combination of individual test images and a customised prompt}
    \label{FIG:Methodology}
\end{figure}

The Figure \ref{FIG:Methodology} gives the basic structure of the experiment. After having downloaded the model into our local system, the test images were passed one after the other and a customised prompt was given to the LMM to predict the class of the image.The experiment's fundamental approach involved determining class labels through a hybrid process, which combined individual test images with a tailored prompting mechanism. Table \ref{TBL:Prompts} shows the different prompts according to the dataset used.

\begin{table} [htb]
\caption{Customised prompts according to dataset}
\begin{center}
\begin{tabular}{|p{0.75in}|p{2in}|}\hline

\textbf{Dataset}&\textbf{Prompt}\\
\hline
MNIST&What number is depicted in the image, choose from 0, 1, 2, 3, 4, 5, 6, 7, 8, 9. Only give the number as the answer, nothing else\\
\hline
Hymnoptera (Ants Vs Bees)&If the image contains ants answer is 0. If the image contains bees answer is 1. Give only the number, nothing else\\
\hline
Cats Vs Dogs&If the image contains cats answer is 0. If the image contains dogs answer is 1. Give only the number, nothing else\\
\hline
Pox Vs No Pox&If the image contains skin with any kind of pox answer is 1. If the image contains skin looking normal answer is 0. Give only the number, nothing else\\
\hline
\end{tabular}
\end{center}
\label{TBL:Prompts}
\end{table}

\subsection{Memory Management}
As the LLaVA 1.5 checkpoint comprises 7 billion parameters, it ordinarily demands approximately 8 GB of GPU resources. Nevertheless, in our experimental setup, we employed a 4-bit quantized variant that operates efficiently on approximately 6 GB of GPU memory. Given that each new image serves as a unique conversational context for the model, reseting the model execution process is necessary. To achieve this, it is imperative to clear both the GPU and RAM before loading a model and introducing a new image to it.

\subsection{System specifications}
In our testing, we utilized an NVIDIA Corporation GP104 GeForce GTX 1070 GPU with 16 GB of dedicated GPU memory, supported by 8 GB of RAM. The operating system was Ubuntu 22.04.1, and we worked with Python version 3.10.

\subsection{Datasets Used}
\begin{figure}[htbp]
    \centering
        \fbox{\includegraphics[scale=.4]{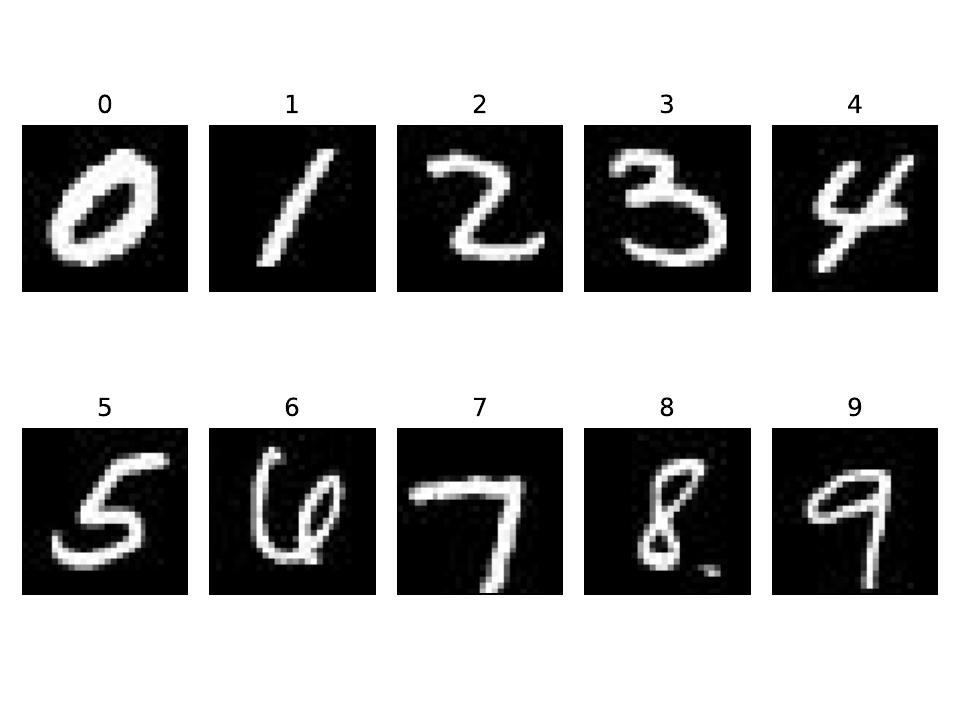}}
    \caption{Images in the MNIST dataset \cite{deng2012mnist}}
    \label{FIG:mnist}
\end{figure}

\begin{figure}[htbp]
    \centering
        \fbox{\includegraphics[scale=.4]{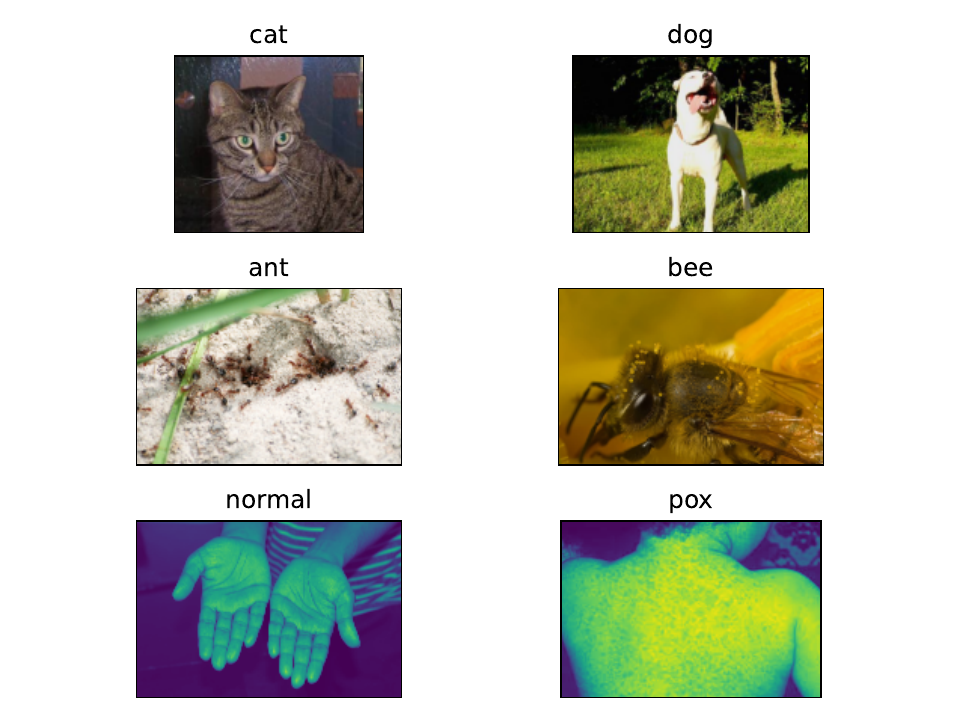}}
    \caption{Images in the CatsVDogs \cite{dogs-vs-cats}, AntsVbees \cite{Melody} and PoxVNoPox \cite{ahsan2022monkeypox} data respectively}
    \label{FIG:others}
\end{figure}

The model was evaluated using the following datasets:

\begin{itemize}
    \item MNIST dataset \cite{deng2012mnist}, which consists of hand-written images representing numbers from 0 to 9. Figure \ref{FIG:mnist} shows the different handwritten images.
    \item The CatsVDogs dataset \cite{dogs-vs-cats}, containing images of various cat and dog breeds. The first row of Figure \ref{FIG:others} shows samples of the images belonging to this dataset.
    \item The Hymnoptera dataset \cite{Melody}, comprising images of ants and bees. In Figure \ref{FIG:others}, the second row displays examples of images from this dataset.
    \item Lastly, the monkeypox dataset, as referenced in \cite{ahsan2022image,ahsan2022monkeypox}, includes images of skin affected by monkeypox, measles, chickenpox, and normal skin. To simplify, images displaying any type of pox were categorized as "pox images," while those with normal skin were grouped in the "normal" category. Samples of images from this dataset are depicted in the third row of Figure \ref{FIG:others}.
\end{itemize}

\subsection{Dataset used for fine-tuning}

\begin{figure}[htbp]
    \centering
        \fbox{\includegraphics[scale=.35]{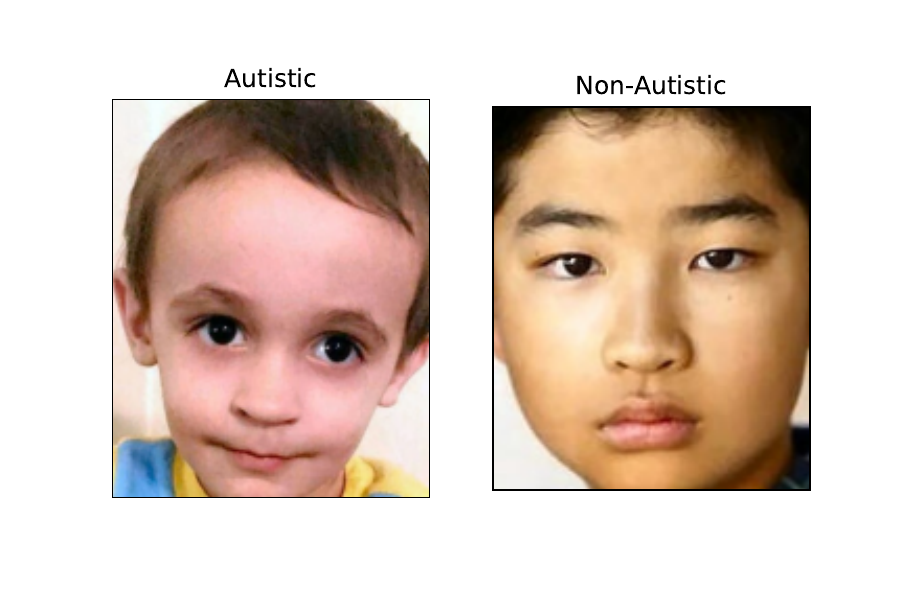}}
    \caption{Images of faces of children with or without autism \cite{Autistic}.}
    \label{FIG:autVnonAut}
\end{figure}

As the second part of the experiment, we have used the Autistic Children Facial Image Data Set \cite{Autistic}. We have selected 200 images from each of autistic and non-autistic class and fine-tuned our model on them. Figure \ref{FIG:autVnonAut} gives an example of two images of faces which are autistic and non-autistic from left to right respectively. 

\begin{figure}[htbp]
    \centering
        \fbox{\includegraphics[scale=.35]{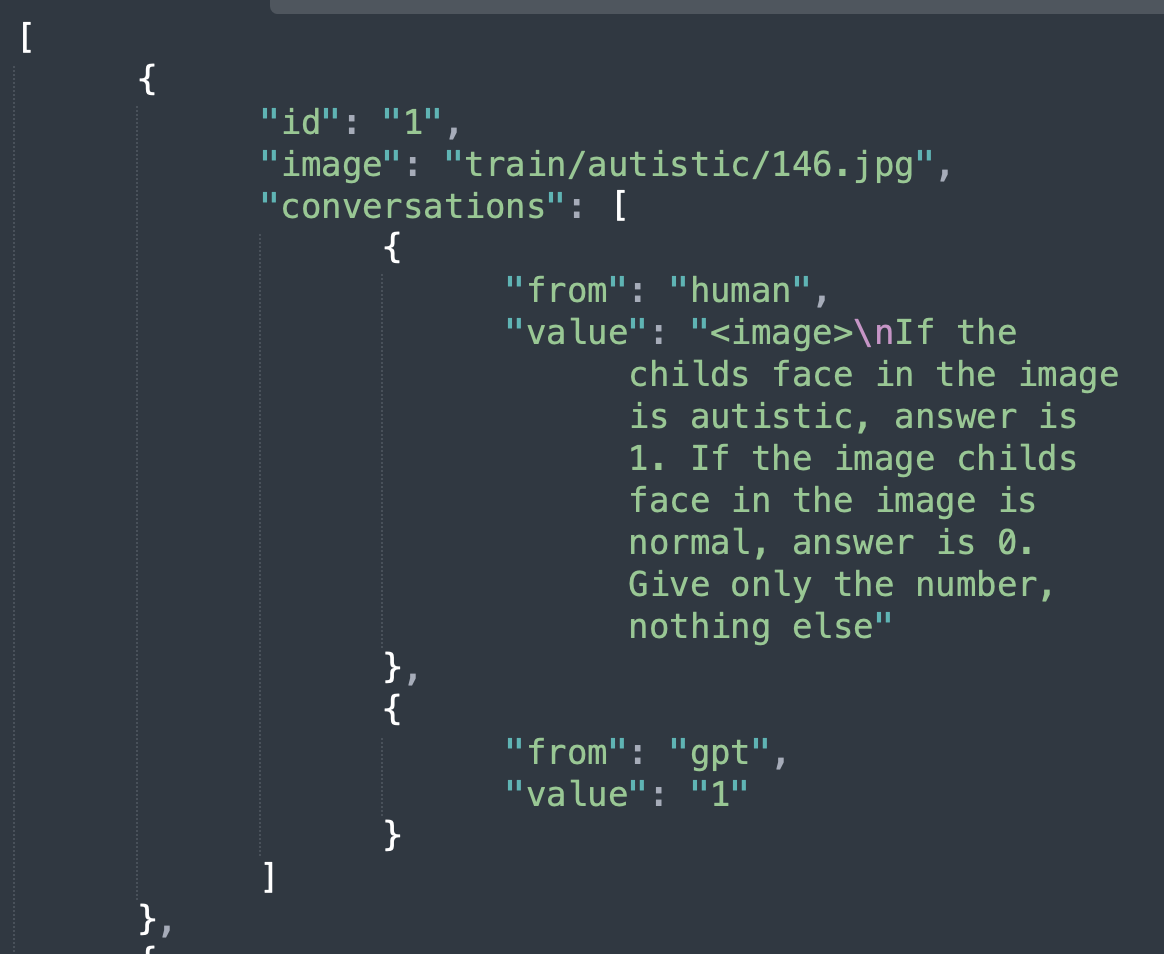}}
    \caption{Json file format to provide necessary prompts to fine-tune the model.}
    \label{FIG:JSONFile}
\end{figure}

Notably, while LLAVA 1.5 exhibits proficiency in facial detection within images, it lacks specialized training to discern whether a facial image displays signs indicative of autism. In fact the zero shot model accuracy on the test dataset of these images is only 55\%, hence the need for fine-tuning.  Our task is to fine tune the model to detect signs of autism in the faces present in these images. The contents of the prompt training file are depicted in Figure \ref{FIG:JSONFile}. Each component within the conversation is represented as a JSON structure, comprising two essential parts: the user's provided prompt and the anticipated model response. Additionally, the JSON structure includes a unique identification number and the filename corresponding to the image that forms the basis of the conversation. The illustrated Figure \ref{FIG:JSONFile} specifically showcases the JSON data corresponding to discussions centered around two distinct image files.

\section{Results}
The model's zero-shot performance has yielded highly promising results, demonstrating a remarkable level of accuracy. Notably, it excels when presented with images featuring large and easily distinguishable objects. As indicated in Table \ref{TBL:Results}, the model achieves a perfect 100\% accuracy on the Cats Vs Dogs dataset, which consists of various cat and dog breeds. Furthermore, it maintains a high level of accuracy on the MNIST dataset, even though the images in this dataset are relatively small, with dimensions of just 28 x 28 pixels.

Our experiments have also uncovered the model's potential applicability to medical datasets. We infer that, when fine-tuned for specific medical image classification tasks, the model can achieve even higher levels of accuracy. This adaptability suggests that the model holds promise for a wide range of applications beyond traditional image classification, extending its utility to the field of medical imaging and diagnosis.

\begin{table} [htb]
\caption{Zero shot Accuracy of LLaVA1.5 on benchmark datasets}
\begin{center}
\begin{tabular}{|p{1.5in}|p{0.5in}|}\hline

\textbf{Dataset}&\textbf{Accuracy}\\
\hline
MNIST&85\%\\
\hline
Hymnoptera (Ants Vs Bees)&77\%\\
\hline
Cats Vs Dogs&100\%\\
\hline
Pox Vs No Pox&79\%\\
\hline
\end{tabular}
\end{center}
\label{TBL:Results}
\end{table}

\subsection{Results On Fine-Tuning}

We completed the fine-tuning of the LLAVA1.5 model on the Autism Face Image dataset. This process was performed on a Google Colab Notebook having A100 (40960MiB) GPU and 89.6 GB RAM. The model fine tuned was the llava-v1.5-7b \cite{liu2023improved, liu2023visual} version for 5 epochs. Other training parameters are as mentioned in the Github repository of the authors of the model \cite{haotian-liu-no-date}. The model was then tested on a balanced test dataset of 50 images from the autistic class and 50 from non-autistic class. These images were not present in the training set. The prompt passed was as below

``\textit{If the childs face in the image is autistic, answer is 1. If the image childs face in the image is normal, answer is 0. Give only the number, nothing else}``

Note that the prompt is same as the prompt used for fine-tuning the model (Figure \ref{FIG:JSONFile}). The results are shown in Table \ref{TBL:ResultsFineTune}.

\begin{table} [htb]
\caption{Accuracy of LLaVA1.5 on the Autism Face Image Dataset before and after fine-tuning}
\begin{center}
\begin{tabular}{|p{1in}|p{1in}|}\hline

\textbf{Before Fine-Tuning}&\textbf{After Fine-Tuning}\\
\hline
55\%&83\%\\
\hline
\end{tabular}
\end{center}
\label{TBL:ResultsFineTune}
\end{table}

A notable increase of nearly 30\% in accuracy is observed following the fine-tuning process for 5 epochs, leveraging a modest dataset of only 400 images. We believe that further optimization of parameters such as the number of images, epochs, and other fine-tuning parameters has the potential to substantially improve the model's accuracy. Such enhancements have the prospect of offering considerable benefits to the medical sciences community.

\subsection{Conclusion}

While LLaVA-1.5 has shown promise in various aspects, it's important to acknowledge the limitations associated with this model:
\begin{itemize}
    \item LLaVA employs full image patches, potentially extending training iterations, with current visual resamplers unable to match its efficiency due to differences in trainable parameters \cite{liu2023improved}.
    \item The model currently lacks the ability to process multiple images, limited by the available instruction-following data and context length \cite{liu2023improved}.
    % \item Proficiency in complex instruction-following notwithstanding, LLaVA-1.5's problem-solving capabilities may be constrained in specific domains, necessitating enhancements via an improved language model and targeted visual instruction tuning data \cite{liu2023improved}.
    \item Despite reduced hallucination tendencies, LLaVA still has the potential to produce hallucinations and misinformation, mandating cautious use in critical applications \cite{liu2023improved}.
\end{itemize}

In spite of these limitations, the achievements of LLaVA-1.5 show the extraordinary potential of multimodal models in the realm of visual reasoning and instruction-following tasks. Our experiments clearly demonstrate its notable accomplishments in zero-shot classification, charting a promising course for future research and innovation. With fine-tuning, the model shows even greater promise to be effective in different critical domains. While researchers work to mitigate its limitations, LLaVA-1.5 remains a guiding light of progress, providing invaluable insights and an easily reproducible framework to advance the frontiers of multimodal AI and elevate its practical utility.

\section*{Acknowledgment}

This publication was partially funded by NPRP grants  14C-0916-210015 and NPRP13S-0206-200281 from the Qatar National Research Fund (a member of Qatar Foundation). The findings herein are solely the responsibility of the authors. The authors would like to also thank Sahar Faramarzi for assisting with the graphics. 

\bibliographystyle{ieeetr}
\bibliography{bibl}

\end{document}